\documentclass[letterpaper]{article} % DO NOT CHANGE THIS
\usepackage{aaai24}  % DO NOT CHANGE THIS
\usepackage{times}  % DO NOT CHANGE THIS
\usepackage{helvet}  % DO NOT CHANGE THIS
\usepackage{courier}  % DO NOT CHANGE THIS
\usepackage[hyphens]{url}  % DO NOT CHANGE THIS
\usepackage{graphicx} % DO NOT CHANGE THIS
\usepackage{natbib}  % DO NOT CHANGE THIS AND DO NOT ADD ANY OPTIONS TO IT
\usepackage{caption} % DO NOT CHANGE THIS AND DO NOT ADD ANY OPTIONS TO IT
\usepackage{algorithm}
\usepackage{algorithmic}
\usepackage{threeparttable}

\usepackage{amsmath}
\usepackage{amssymb}
\usepackage{multirow}
\usepackage{tabularx} % for 'tabularx' env. and 'X' col. type
\usepackage{booktabs} % for \toprule, \midrule etc macros
\usepackage{marvosym}
\usepackage{newfloat}
\usepackage{listings}
\DeclareCaptionStyle{ruled}{labelfont=normalfont,labelsep=colon,strut=off} % DO NOT CHANGE THIS
\floatstyle{ruled}
\newfloat{listing}{tb}{lst}{}
\floatname{listing}{Listing}
\title{SkeletonGait: Gait Recognition Using Skeleton Maps}
\author {
    Chao Fan\textsuperscript{\rm 1,2},
    Jingzhe Ma\textsuperscript{\rm 1,2},
    Dongyang Jin\textsuperscript{\rm 1,2}, 
    Chuanfu Shen\textsuperscript{\rm 1,2,3}, 
    Shiqi Yu\textsuperscript{\rm 1,2 \Letter}
}
\affiliations {
    \textsuperscript{\rm 1}Research Institute of Trustworthy Autonomous System, Southern University of Science and Technology \\ 
    \textsuperscript{\rm 2}Department of Computer Science and Engineering, Southern University of Science and Technology \\
    \textsuperscript{\rm 3}The University of Hong Kong \\
    \{12131100, 12031127, 11911221, 11950016\}@mail.sustech.edu.cn, yusq@sustech.edu.cn
}

\usepackage{bibentry}
\begin{document}

\maketitle

\begin{abstract}
The choice of the representations is essential for deep gait recognition methods. The binary silhouettes and skeletal coordinates are two dominant representations in recent literature, achieving remarkable advances in many scenarios. However, inherent challenges remain, in which silhouettes are not always guaranteed in unconstrained scenes, and structural cues have not been fully utilized from skeletons. 
In this paper, we introduce a novel skeletal gait representation named skeleton map, together with SkeletonGait, a skeleton-based method to exploit structural information from human skeleton maps. Specifically, the skeleton map represents the coordinates of human joints as a heatmap with Gaussian approximation, exhibiting a silhouette-like image devoid of exact body structure. 
Beyond achieving state-of-the-art performances over five popular gait datasets, more importantly, SkeletonGait uncovers novel insights about how important structural features are in describing gait and when they play a role. 
Furthermore, we propose a multi-branch architecture, named SkeletonGait++, to make use of complementary features from both skeletons and silhouettes. 
Experiments indicate that SkeletonGait++ outperforms existing state-of-the-art methods by a significant margin in various scenarios.
For instance, it achieves an impressive rank-1 accuracy of over $85\%$ on the challenging GREW dataset.
All the source code is available at \url{https://github.com/ShiqiYu/OpenGait}.
\end{abstract}

\section{1.~Introduction}
Vision-based gait recognition refers to the use of vision technologies for individual identification based on human walking patterns. 
Compared to other biometric techniques such as face, fingerprint, and iris recognition, gait recognition offers the benefits of non-intrusive and long-distance identification without requiring the cooperation of the subject of interest. 
These advantages make gait recognition particularly suitable for various security scenarios such as suspect tracking and crime investigation~\cite{nixon2006automatic}.

\begin{figure}[t]
\centering
\includegraphics[height=7.5cm]{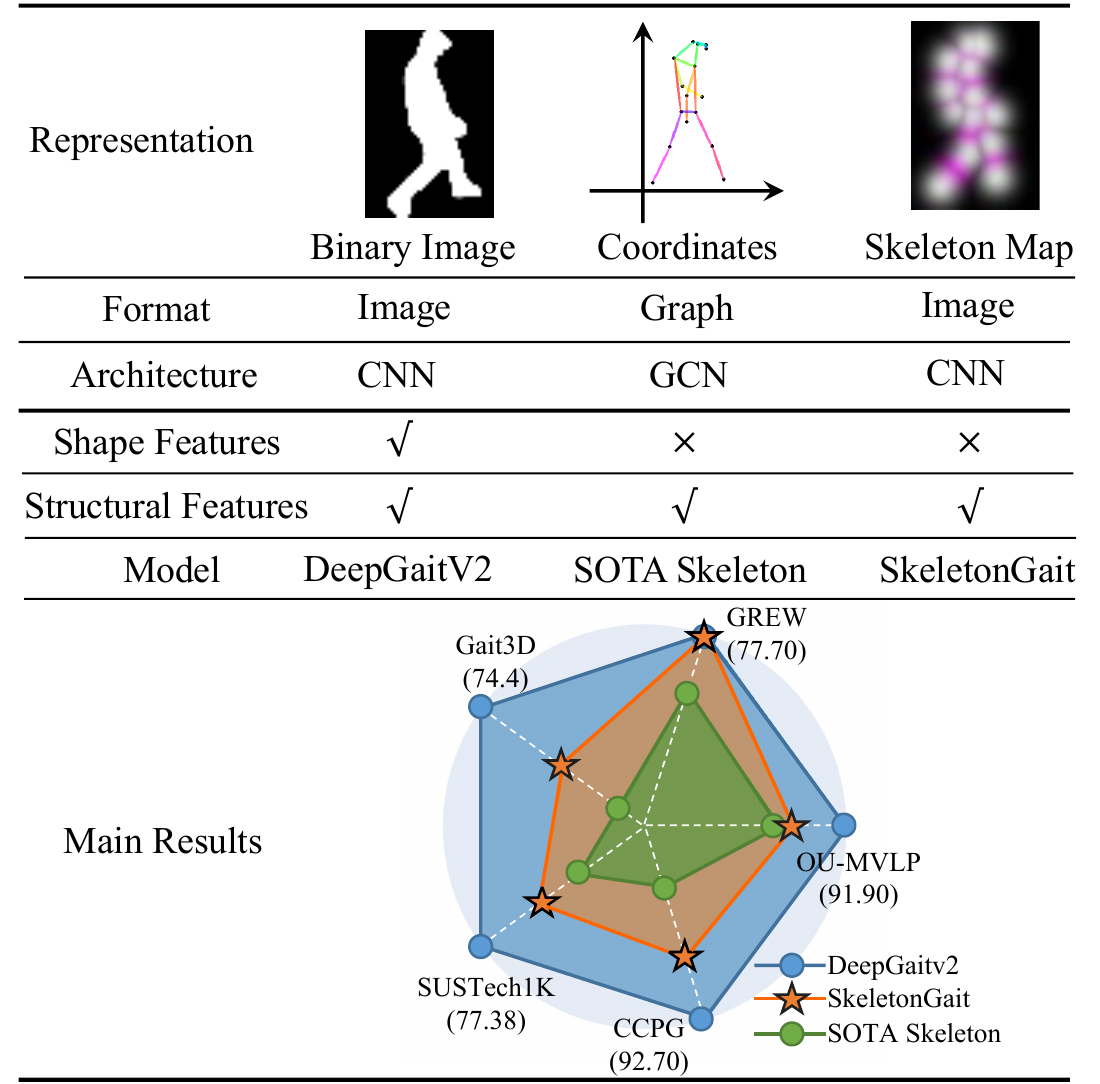}
\caption{The representations of the developed skeleton map \textit{v.s.} the classical gait graph and silhouette. Only a single frame is displayed for brevity. }
\label{fig:intro}
\end{figure}

Before leveraging deep models to learn gait features, a fundamental issue worth exploring is to consider the ideal input modality. 
To achieve robust long-term human identification, this input should be the `clean' gait representation maintaining gait-related features such as body shape, structure, and dynamics, and meanwhile eliminate the influence of gait-unrelated factors, such as background, clothing, and viewpoints. 
In recent literature, the binary silhouettes and skeletons serve as the two most prevailing gait representations~\cite{shen2022comprehensive}. 
As shown in Fig.~\ref{fig:intro}, they both explicitly present the structural characteristics of the human body, \textit{e.g.}, the length, ratio, and movement of human limbs.
Silhouettes, differently, have more discriminative capacity by explicitly maintaining appearance information. 
However, utilizing appearance information from silhouettes is not always beneficial for identification, as these characteristics are usually vulnerable and mixed up with the shape of dressing and carrying items. 
Conversely, skeletons present an appearance-free representation and are naturally robust to appearance changes. 
Nevertheless, existing skeleton-based methods primarily employ Graph Convolutional Networks (GCNs) on conventional skeletal representations (\textit{i.e.} 2D/3D coordinates) and provide unsatisfactory performance, particularly with real-world applications. 

To explore the cooperativeness and complementarity natures of body shape and structural features, this paper introduces a novel skeleton-based gait representation called \textbf{Skeleton Map}, drawing inspirations from related works~\cite{duan2022revisiting, liu2018recognizing, liao2022posemapgait}. 
As illustrated in Fig.~\ref{fig:intro}, the skeleton map represents the coordinates of human joints as a heatmap with Gaussian approximation and gait-oriented designs. 
This approach aligns skeleton and silhouette data across spatial-temporal dimensions, representing the skeleton as a silhouette-like image without exact body shapes.
To further align the network architectures, we introduce a baseline model referred to as \textbf{SkeletonGait}.
This model is developed by replacing the input of DeepGaitV2~\cite{fan2023exploring} from the conventional silhouette to the skeleton map. 
This straightforward design is strongly motivated by two-fold considerations: 
a) We establish the alignments between SkeletonGait and DeepGaitV2 in terms of both input data format and network architectures, facilitating an intuitive comparison of the representational capacities of solely body structural features \textit{v.s.} the combination of body shape and structural features\footnote{
We consider the primary difference between the silhouette and skeleton is their inclusion or exclusion of the body shape. 
The body shape removal can effectively eliminate self-occlusions. Therefore, this paper views self-occlusion as a passenger variable brought by shape removal, thus not directly serving as a causal factor.
}.
b) Notably, DeepGaitV2 has achieved the latest state-of-the-art performance on various gait datasets, motivating the adoption of its architecture as the baseline for this paper.

As shown in Fig.~\ref{fig:intro}, we present a comprehensive evaluation on five popular large-scale gait datasets: OU-MVLP~\cite{takemura2018multi}, GREW~\cite{zhu2021gait}, Gait3D~\cite{zheng2022gait}, SUSTech1K~\cite{Shen_2023_CVPR}, and CCPG~\cite{Li_2023_CVPR}.
Here the label  `SOTA Skeleton' denotes the most cutting-edge performances achieved by existing skeleton-based methods, regardless of the sources of publication. 
According to in-depth investigations, we have uncovered the following insights:
1) Compared with previous skeleton-based methods, SkeletonGait better exposes the importance of body structural features in describing gait patterns thanks to its competitiveness. 
The underlying reasons, \textit{i.e.}, the advantages of the skeleton map over raw joint coordinates, will be carefully discussed.
2) Interestingly, despite GREW is usually regarded as the most challenging gait dataset due to its extensive scale and real-world settings, SkeletonGait performing impressive performance suggests that the walking patterns of its subjects can be effectively represented solely by body structural attributes, with no requirement for shape characteristics. 
This revelation prompts a subsequent investigation into the potential lack of viewpoint diversity of GREW.
3) When the input silhouettes become relatively unreliable, such as in instances of poor illumination in SUSTech1K and complex occlusion in Gait3D and GREW, the skeleton map emerges as a pivotal player in discriminative and robust gait feature learning. 
Further findings and insights will be discussed in the experiment section. 

By integrating the superiority of silhouette and skeleton map, a novel gait framework known as \textbf{SkeletonGait++} is introduced.
In practice, SkeletonGait++ effectively aggregates the strengths of these two representations by a fusion-based multi-branch architecture.
Experiments show that SkeletonGait++ reaches a pioneering state-of-the-art performance, surpassing existing methods by a substantial margin. 
Further visualizations verify that SkeletonGait++ is capable of adaptively capturing meaningful gait patterns, consisting of discriminative semantics within both body structural and shape features.

Overall, this paper promotes gait research in three aspects: 
\begin{itemize}
    \item The introduction of the skeleton map aligns two widely employed gait representations, namely the skeleton and silhouette, in terms of input data format. This alignment facilitates an intuitive exploration of their collaborative and complementary characteristics.
    \item SkeletonGait introduces a robust baseline model utilizing skeleton maps, showcasing remarkable advancements over preceding skeleton-based methods across diverse gait datasets. Beyond its quantitative achievements, the insights and revelations derived from SkeletonGait can inspire further gait research.
    \item The multi-modal SkeletonGait++ reaches a new state-of-the-art across various datasets by extracting `comprehensive' gait features. 
\end{itemize}

\section{2.~Related Works}
\textbf{Gait Representations.} The popular gait representations are primarily derived from RGB images, including raw RGB images, binary silhouettes, optical images, 2D/3D skeletons, and human meshes. 
To mitigate the influence of extraneous noise stemming from color, texture, and background elements, these representations often rely on preprocessing stages or end-to-end learning approaches.
Beyond the typical RGB cameras, some studies propose novel gait representations by incorporating emerging sensors such as LiDAR~\cite{Shen_2023_CVPR} and event cameras~\cite{9337225}. 
However, these sensors are currently less commonly found in existing CCTVs, making them temporarily unsuitable for large-scale video surveillance applications. 
This paper focuses on two of the most widely-used gait representations, \textit{i.e.} silhouette and skeleton data. 

According to the classical taxonomy, gait recognition methods can be broadly classified into two categories: model-based and appearance-based methods.

\noindent
\textbf{Model-based Gait Recognition} methods utilize the underlying structure of the human body as input, such as the estimated 2D / 3D skeleton and human mesh. 
With extremely excluding visual clues, these gait representations, which are formally parameterized as coordinates of human joints or customized vectors in most cases, are theoretically `clean' against factors like carrying and dressing items. 
In recent literature, 
PoseGait~\cite{liao2020model} combines the 3D skeleton data with hand-crafted characteristics to overcome the viewpoint and clothing variations, GaitGraph~\cite{teepe2021gaitgraph} introduces a graph convolution network for 2D skeleton-based gait representation learning, 
HMRGait~\cite{li2020end} fine-tunes a pre-trained human mesh recovery network to construct an end-to-end SMPL-based model, 
Despite the advances achieved on indoor OU-MVLP, previous model-based methods still have not exhibited competitive performance compared with the appearance-based ones on real-world gait datasets. 

\noindent
\textbf{Appearance-based Gait Recognition} methods mostly learn gait features from silhouette or RGB images, leveraging informative visual characteristics. 
With the advent of deep learning, current appearance-based approaches primarily concentrate on spatial feature extraction and gait temporal modeling.
Specifically, GaitSet~\cite{chao2019gaitset} innovatively treats the gait sequence as a set and employs a maximum function to compress the sequence of frame-level spatial features. 
Due to its simplicity and effectiveness, GaitSet has emerged as one of the most influential gait recognition works in recent years.
GaitPart~\cite{fan2020gaitpart} meticulously explores the local details of input silhouettes and models temporal dependencies using the Micro-motion Capture Module.
GaitGL~\cite{lin2021gait} argues that spatially global gait representations often overlook important details, while local region-based descriptors fail to capture relationships among neighboring parts. Consequently, GaitGL introduces global and local convolution layers.
More recently, DeepGaitV2~\cite{fan2023exploring} presents a unified perspective to explore how to construct deep models for outdoor gait recognition, bringing a breakthrough improvement on the challenging Gait3D and GREW.

Additionally, there are also some progressive multi-modal gait frameworks, such as SMPLGait~\cite{zheng2022gait} that exploited the 3D geometrical information from the SMPL model to enhance the gait appearance feature learning, and BiFusion~\cite{peng2023learning} that integrated skeletons and silhouettes to capture the rich gait spatiotemporal features.

\noindent
\textbf{Related Works to Skeleton Map.}
Liu \textit{et al.}\cite{liu2018recognizing} introduced the aggregation of pose estimation maps, which are intermediate feature maps from skeleton estimators, to create a heatmap-based representation for action recognition. 
This idea has been extended to gait recognition by Liao \textit{et al.}\cite{liao2022posemapgait}. 
However, the intermediate feature often involves float-encoded noises, potentially incorporating body shape information that is undesirable for model-based gait applications. 
Additionally, Liao \textit{et al.}~\cite{liao2022posemapgait} have not demonstrated competitive results on the challenging outdoor gait datasets using pose heatmaps.
Similar to the approach in~\cite{duan2022revisiting}, our skeleton map is generated solely from the coordinates of human joints, deliberately excluding any potential visual clues hidden in pose estimation maps. 
But differently, we place emphasis on the pre-treatment of data and the design of deep models for gait recognition purposes.

\section{3.~Method}
This section begins with outlining the generation of skeleton maps. 
Subsequently, we delve into the specifics of SkeletonGait and SkeletonGait++.
Implementation details are introduced at the end of this section. 

\begin{figure}[t]
\centering
\includegraphics[height=3.2cm]{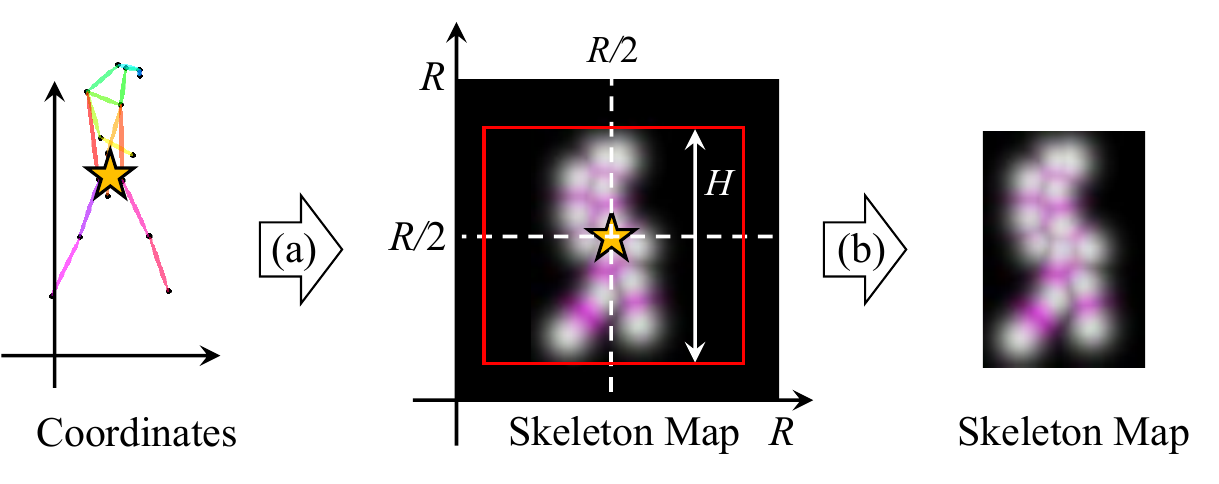}
\caption{The pipeline of skeleton map generation. (a) Center-normalization, scale-normalization, and skeleton rendering. (b) Subject-centered cropping.}
\label{fig:pipeline}
\end{figure}

\subsection{3.1~Skeleton Map}
Given the coordinates of human joints $(x_k, y_k, c_k)$, where $(x_k, y_k)$ and $c_k$ respectively present the location and confidence score of the $k$-th joint with $k\in\{1, ..., K\}$, we generate the skeleton map by following steps.

Firstly, considering the absolute coordinates of joints relative to the original image contain much gait-unrelated information like the walking trajectory and filming distance, we introduce the pre-treatments of center- and scale-normalization to align raw coordinates: 
\begin{equation}
    \begin{aligned}
        x_k &= x_k - x_{\text{core}} + R/2 \\ 
        y_k &= y_k - y_{\text{core}} + R/2 \\ 
        x_k &= \frac{x_k - y_{\text{min}}}{y_{\text{max}} - y_{\text{min}}} \times H\\ 
        y_k &= \frac{y_k - y_{\text{min}}}{y_{\text{max}} - y_{\text{min}}} \times H
    \end{aligned}
\label{equ:normalization}
\end{equation}
where $(x_{\text{core}}, y_{\text{core}})=(\frac{x_{11} + x_{12}}{2}, \frac{y_{11} + y_{12}}{2})$ presents the center point of two hips (11-th and 12-th human joints, their center can be regarded as the barycenter of the human body),
and $(y_{\text{max}}, y_{\text{min}})$ denotes the maximum and minimum heights of human joints $(\underset{k}{\text{max}}y_k , \underset{k}{\text{min}}y_k)$. 
In this way, we move the barycenter of the human body to $(R/2, R/2)$ and normalize the body height to $H$, as shown in Fig.~\ref{fig:pipeline}(a).

Typically, the height of the human body is expected to exceed its width.
As a result, the normalized coordinates of human joints, as defined in Eq.~\ref{equ:normalization}, should fall within the range of $H \times H$. 
But in practice, the pose estimator is imperfect and may produce some outlier joints outside the $H\times H$ scope. 
To address these out-of-range cases, the resolution of the skeleton map, denoted as $R$, should be larger than $H$, ensuring coverage of all the coordinates. 
In our experiments, let $R$ be $2H$ is enough for the OUMVLP, GREW, Gait3D, CCPG, and SUSTech1K datasets. 

\begin{figure}[t]
\centering
\includegraphics[height=7.5cm]{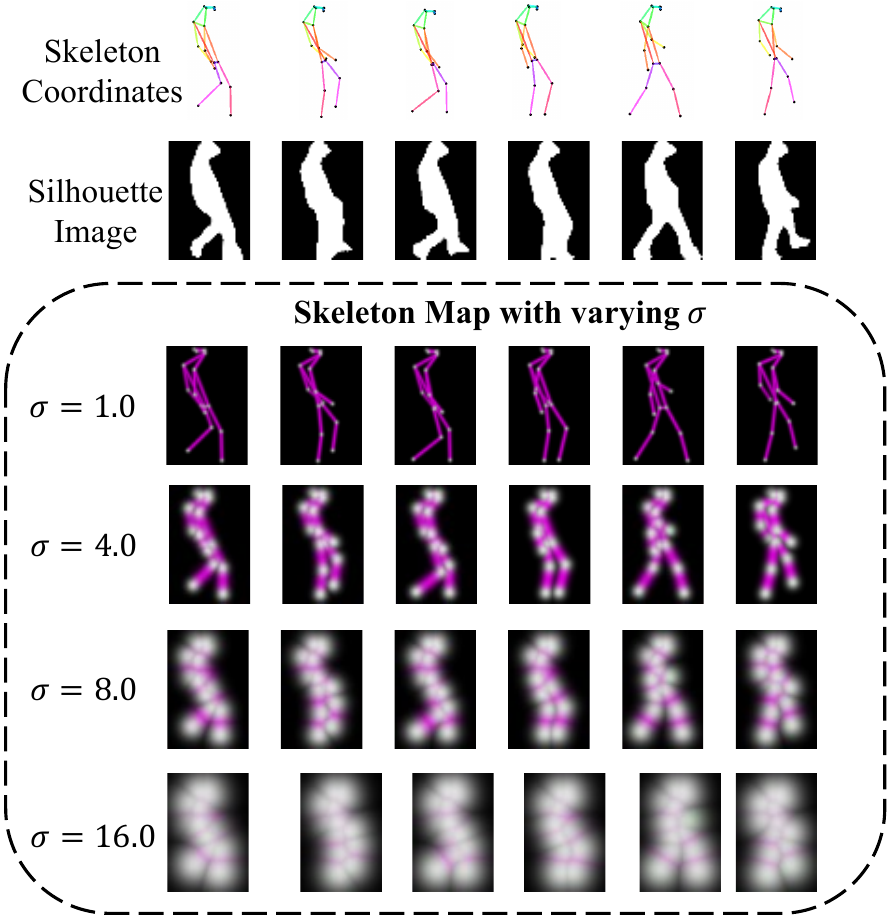}
\caption{More examples of the skeleton coordinates \textit{v.s.} silhouette images \textit{v.s.} skeleton maps.}
\label{fig:examples}
\end{figure}

As illustrated in Figure~\ref{fig:pipeline} (a), the skeleton map is initialized as a blank image with a size of $R \times R$.
Then we draw it based on the normalized coordinates of human joints.
Inspired by~\cite{duan2022revisiting}, we generate the joint map $\textit{\textbf{J}}$ by composing $K$ Gaussian maps, where each Gaussian map is centered at a specific joint position and contributes to all the $R \times R$ pixels: 
\begin{equation}
    \textit{\textbf{J}}_{(i, j)}=\sum_{k}^{K}e^{-\frac{(i - x_k)^2 + (j - y_k)^2}{2\sigma^2}} \times c_k
\label{equ:joints}
\end{equation}
where $\textit{\textbf{J}}_{(i, j)}$ presents the value of a certain point from $\{(i, j)|i,j\in \{1, ..., R\}\}$, and $\sigma$ is a hyper-parameter controlling the variance of Gaussian maps. 

Similarly, we can also create a limb map $\textit{\textbf{L}}$: 
\begin{equation}
    \textit{\textbf{L}}_{(i, j)}=\sum_{n}^{N}e^{-\frac{\mathcal{D}((i,j), \mathcal{S}[n^-, n^+])^2}{2\sigma^2}} \times \text{min}(c_{n^-}, c_{n^+})
\label{equ:limbs}
\end{equation}
where $\mathcal{S}[n^-, n^+]$ presents the $n$-th limb determined by $n^-$-th and $n^+$-th joints with $n^-, n^+\in \{1, ..., K\}$. 
The function $\mathcal{D}((i,j), \mathcal{S}[n^-, n^+])$ measures the Euclidean distance from the point $(i, j)$ to the $n$-th limb, where $n\in\{1, ..., N\}$ and $N$ denotes the count of limbs. 
% with the coordinates-triplets of $(x_{n^-}, y_{n^-}, c_{n^-})$ and $(x_{n^+}, y_{n^+}, c_{n^+})$. 

Next, the skeleton map is obtained by stacking $\textit{\textbf{J}}$ and $\textit{\textbf{L}}$ and thus has a size of $2\times R \times R$.
Notably, for the convenience of visualization, we repeat the last channel of all the skeleton maps shown in this paper to display the visual three-channel images with the size of $3\times R \times R$. 

As shown in Figure~\ref{fig:pipeline} (b), we employ a subject-centered cropping operation to remove the unnecessary blank regions, thus reducing the redundancy in skeleton maps. 
In practice, the vertical range is determined by the minimum and maximum heights of pixels which possess non-zero values. 
Meanwhile, the horizontal cropping range spans from $\frac{R-H}{2}$ to $\frac{R+H}{2}$. 
In this way, we remove extraneous areas outside the desired gait region, ensuring a more concise and compact skeleton map.
Lastly, to align with the input size required by downstream gait models, the cropped skeleton maps are resized to $2\times 64 \times 64$ and further cropped by the widely-used double-side cutting strategy. 

As a result, Fig.~\ref{fig:examples} exhibits some examples of the used skeleton maps with varying $\sigma$.
As we can see, a smaller $\sigma$ produces a visually thinner skeleton map, whereas excessively large $\sigma$ may lead to visual ambiguity. 

Compared with approaches proposed by~\cite{duan2022revisiting, liu2018recognizing, liao2022posemapgait}, 
our skeleton map introduces the following gait-oriented enhancements: 
\begin{itemize}
    % \item \textbf{Improved Representational Ability}. 
    % As shown in Figure~\ref{fig:pipeline}, our skeleton map effectively captures the body structural characteristics like the length, ratio, and movements of limbs, which are essential for the downstream spatiotemporal gait representation learning, thus being well-suited for gait recognition tasks. 
    \item 
    \textbf{Cleanness}. The implementation of center-normalization effectively eliminates identity-unrelated noise present in raw skeleton coordinates, \textit{i.e.}, the walking trajectory, and camera distance information.
    \item 
    \textbf{Discriminability}. Preceding methods tend to directly resize the obtained images of varying sizes into a predetermined fixed size, inevitably resulting in the loss of body ratio information. Conversely, the scale-normalization and subject-centered cropping techniques outlined in this paper ensure that the skeleton map preserves the authenticity of the length and ratio of human limbs.
    \item 
    \textbf{Compactness}. All the joints and limbs are drawn within a single map, optimizing the efficiency of the modeling process, as opposed to a stack of separate maps.
\end{itemize}

\begin{figure*}[t]
\centering
\includegraphics[height=3.5cm]{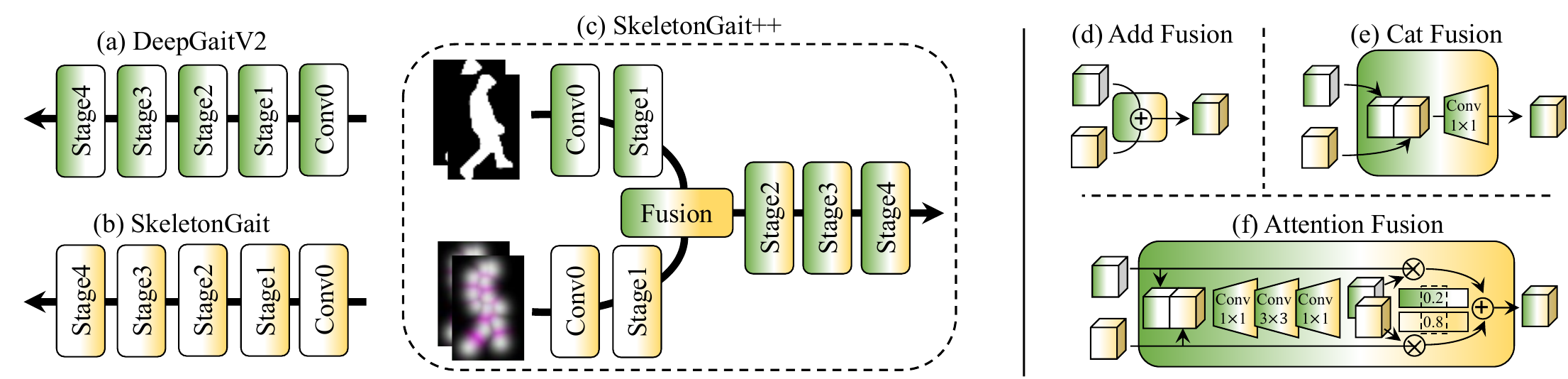}
\caption{The network architectures of DeepGaitV2 \textit{v.s.} SkeletonGait \textit{v.s.} SkeletonGait++. The `head' part is ignored for brevity.}
\label{fig:skeletongait}
\end{figure*}

Previous skeleton-based gait recognition methods tend to model the coordinates of joints as non-grid gait graphs with learnable edges, potentially losing inherent structural priors within a highly structured human body. 
In this paper, the proposed skeleton map is a kind of grid-based skeletal gait representation, where the body structural characteristics highly desired by gait recognition, such as the length, ratio, and movement of body limbs, are explicitly and naturally distributed over the spatial and temporal dimensions, exactly matching the locality modeling requirement of fine-grained spatiotemporal gait description. 
Moreover, the skeleton map offers additional advantages:
\begin{itemize}
    \item The skeleton map shares similarities with gait graphs in terms of feature content and with silhouettes in terms of data format. This unique characteristic allows the skeleton map to benefit from recent advancements in both skeleton-based and silhouette-based methods. 
    \item Interestingly, the skeleton map can be perceived as a silhouette that excludes body shape information, facilitating an intuitive comparison of the representational capacities of solely body structural features \textit{v.s.} the combination of body shape and structural features.
    \item As an imagery input,  the skeleton map can seamlessly integrate into image-based multi-modal gait models, particularly at the bottom stages of the model. 
\end{itemize}

\subsection{3.2~SkeletonGait}
\label{sec:skeletongait}
Ideally, we can employ any image-based gait methods to build a skeleton-map-based baseline model. 
In this paper, SkeletonGait is developed by replacing the input of DeepGaitV2~\cite{fan2023exploring} from the silhouette to skeleton map, as shown in Fig.~\ref{fig:skeletongait}(a) and (b), 
The only architectural modification is to change the input channel of the Conv0, where the silhouette is a single-channel input and the skeleton map is a double-channel input.
This straightforward design is strongly motivated by two primary reasons: 
\begin{itemize}
    \item The alignment of network architectures enables a seamless and intuitive comparative study between the silhouette and skeleton map representations.
    \item DeepGaitV2 has a straightforward architecture providing state-of-the-art performances across various gait datasets, making it well-suited for benchmarking.
\end{itemize}

\subsection{3.3~SkeletonGait++}
To integrate the superiority of silhouette and skeleton map, as shown in Fig.~\ref{fig:skeletongait}(c), 
SkeletonGait++ provides a fusion-based two-branch architecture involving the silhouette and skeleton branches. 
These two branches respectively share the same network architectures with DeepGaitV2 and SkeletonGait at early stages, such as the Conv0 and Stage1.

\begin{table}[tb]
\centering
\caption{Implementation details.
The batch size $(q, k)$ indicates $q$ subjects with $k$ sequences per subject.}
\resizebox{0.46\textwidth}{!}{
\begin{tabular}{cccc}
% \hline
\toprule
 DataSet & Batch Size & Milestones & Total Steps \\ \midrule %\hline
 OUMVLP              & (32, 8)                    & (60k, 80k, 100k)              & 120k                       \\ 
 CCPG                & (8, 16)                    & (20k, 40K, 50k)               & 60k                    \\ 
 SUSTech1K           & (8,  8)                    & (20k, 30k, 40k)               & 50k                    \\
 Gait3D              & (32, 4)                    & (20k, 40K, 50k)               & 60k                    \\ 
 GREW                & (32, 4)                    & (80k, 120k, 150k)             & 180k                    \\
 \bottomrule 
\end{tabular}
}
\label{tab:implementations}
\end{table}

Then, a fusion module is responsible for aggregating these two feature sequences frame-by-frame. 
For the sake of brevity, Fig.~\ref{fig:skeletongait} displays a single frame while ensuring correctness, as frames are processed in parallel. 
In this paper, we consider three kinds of fusion mechanisms: 
\begin{itemize}
    \item \textbf{Add Fusion.} The feature maps from the silhouette and skeleton branch are combined using an element-wise addition operation, as demonstrated in Fig.~\ref{fig:skeletongait}(d).
    \item \textbf{Concatenate Fusion.} The feature maps from the silhouette and skeleton branch are first concatenated along the channel dimension, and then transformed by a plain $1\times1$ convolution layer, as demonstrated in Fig.~\ref{fig:skeletongait}(e). 
    \item \textbf{Attention Fusion.} The feature maps from the silhouette and skeleton branch are first concatenated along the channel dimension, and then transformed by a small network to form a cross-branch understanding. 
    Here the small network is composed of a squeezing $1\times1$, a plain $3\times3$, and an expansion $1\times1$ convolution layer. 
    As shown in Fig.~\ref{fig:skeletongait}(e), a softmax layer is next employed to assign element-wise attention scores respectively for the silhouette and skeleton branch. 
    Lastly, an element-wise weighted-sum operation is used to generate the output.
\end{itemize}

Next, the Stage 3 and 4 possess the same network architectures as the SkeletonGait. 
Moreover, we also consider the fusion location. 
Fig.~\ref{fig:skeletongait}(c) exhibits the \textbf{Low-Level} fusion case.
Another \textbf{High-Level} fusion model aggregates the features before Stage 4, with additional Stage 2 and 3 respectively being inserted into the silhouette and skeleton branch. 

\begin{table}[tb]
\centering
\caption{
Datasets in use. \#ID and \#Seq present the number of identities and sequences.
}
\resizebox{0.5\textwidth}{!}{
\begin{tabular}{cccccc}
\toprule
\multirow{2}{*}{DataSet} & \multicolumn{2}{c}{Train Set} & \multicolumn{2}{c}{Test Set} & \multirow{2}{*}{\begin{tabular}[c]{@{}c@{}}Collection\\ situations\end{tabular}} \\
                         & Id            & Seq           & Id           & Seq           &     \\ \midrule %\hline  
OU-MVLP                  & 5,153         & 144,284       & 5,154        & 144,412       & Constrained                                                                      \\
CCPG                  & 100            & 8,187         & 100           & 8,095         & Constrained                                                              \\
SUSTech1K                  & 200         & 5,988       & 850        & 19,228       & Constrained                                                                      \\
Gait3D                   & 3,000         & 18,940        & 1,000        & 6,369         & Real-world                                                            \\
GREW                     & 20,000        & 102,887       & 6,000        & 24,000        & Real-world                                                            \\
\bottomrule
% \hline
\end{tabular}
}
\label{tab:datasets}
\end{table}

\begin{table*}[tb]
\centering
\caption{Recognition results on three authoritative gait datasets, involving OUMVLP, GREW, and Gait3D. 
% The identical-view cases are excluded for the OUMVLP dataset. 
The best performances are in \underline{\textbf{blod}}, and that by skeleton-based methods are in \textbf{blod}. The same annotation is applied in the following table. }
\resizebox{2.0\columnwidth}{!}{
\begin{threeparttable}
\begin{tabular}{c|c|c|ccccccccc}
% \hline
\toprule
\multirow{3}{*}{Input}      & \multirow{3}{*}{Method} & \multirow{3}{*}{Source} & \multicolumn{9}{c}{Testing Datasets}                                                                                     \\ \cline{4-12} 
                            &                         &                         & \multicolumn{1}{c|}{OU-MVLP} & \multicolumn{4}{c|}{GREW}                                & \multicolumn{4}{c}{Gait3D}     \\ \cline{4-12} 
                            &                         &                         & \multicolumn{1}{c|}{rank-1}  & rank-1 & rank-5 & rank-10 & \multicolumn{1}{c|}{rank-20} & rank-1 & rank-5  & mAP  & mINP \\ \midrule %\hline
\multirow{3}{*}{\begin{tabular}[c]{@{}c@{}} Skeleton  \\ Coordinates \end{tabular}}   & GaitGraph2              & CVPRW2022               & \multicolumn{1}{c|}{62.1}    & 33.5   & \multicolumn{3}{c|}{-}                          & 11.1   & \multicolumn{3}{c}{-} \\
                            & GaitTR                  & Arxiv2022               & \multicolumn{1}{c|}{56.2}    & 54.5   & \multicolumn{3}{c|}{-}                          & 6.6    & \multicolumn{3}{c}{-} \\
                            & GPGait                  & ICCV2023               & \multicolumn{1}{c|}{60.5}    & 53.6   & \multicolumn{3}{c|}{-}                          & 22.5   & \multicolumn{3}{c}{-} \\ \midrule
Skeleton Maps               & SkeletonGait                & Ours                    & \multicolumn{1}{c|}{\textbf{67.4$^\dag$}}    & \textbf{77.4}   & \textbf{87.9}   & \textbf{91.0}    & \multicolumn{1}{c|}{\textbf{93.2}}    & \textbf{38.1}   & \textbf{56.7}    & \textbf{28.9} & \textbf{16.1}   \\ \midrule %\hline
\multirow{5}{*}{Silhouette} & GaitSet                 & AAAI2019                & \multicolumn{1}{c|}{87.1}    & 46.3   & 63.6   & 70.3    & \multicolumn{1}{c|}{-}       & 36.7   & 58.3    & 30.0 & 17.3 \\
                            & GaitPart                & CVPR2020                & \multicolumn{1}{c|}{88.5}    & 44.0   & 60.7   & 67.3    & \multicolumn{1}{c|}{-}       & 28.2   & 47.6    & 21.6 & 12.4 \\
                            & GaitGL                  & ICCV2021                & \multicolumn{1}{c|}{89.7}    & 47.3   & \multicolumn{3}{c|}{-}                          & 29.7   & 48.5    & 22.3 & 13.6 \\
                            & GaitBase                & CVPR2023                & \multicolumn{1}{c|}{90.8}    & 60.1   & \multicolumn{3}{c|}{-}                          & 64.6   & \multicolumn{3}{c}{-} \\
                            & DeepGaitV2          & Arxiv2023               & \multicolumn{1}{c|}{\underline{\textbf{91.9}}}    & 77.7   & 88.9   & 91.8    & \multicolumn{1}{c|}{-}       & 74.4   & 88.0    & 65.8 & -    \\ 
                            \midrule %\hline
\multirow{3}{*}{\begin{tabular}[c]{@{}c@{}}Silhouette+\\ Skeleton / SMPL\end{tabular}} 
& SMPLGait                & CVPR2022                & \multicolumn{1}{c|}{-}       & \multicolumn{4}{c|}{-}                                   & 46.3   & 64.5    & 37.2 & 22.2 \\ 
& GaitRef                & IJCB2023                & \multicolumn{1}{c|}{90.2}   & 53.0   & 67.9   & 73.0 & \multicolumn{1}{c|}{77.5}  & 49.0   & 49.3    & 40.7 & 25.3 \\ \cmidrule{2-12}
                            & SkeletonGait++          & Ours               & \multicolumn{1}{c|}{-$^\ddag$}    & \underline{\textbf{85.8}}   & \underline{\textbf{92.6}}   & \underline{\textbf{94.3}}    & \multicolumn{1}{c|}{\underline{\textbf{95.5}}}       & \underline{\textbf{77.6}}   & \underline{\textbf{89.4}}    & \underline{\textbf{70.3}} & \underline{\textbf{42.6}}    \\ \bottomrule %\hline
\end{tabular}
\textit{$\dag$ For OU-MVLP, we conducted experiments using both AlphaPose and OpenPose data, resulting in rank-1 accuracy of 67.4\% and 65.9\%, respectively. These results consistently surpass other pose-based methods, revealing the robustness of SkeletonGait to different pose estimators.} \\
\textit{$\ddag$ The lack of results for SkeletonGait++ on OU-MVLP is due to the absence of frame-by-frame alignment between the skeleton and silhouette.}
\end{threeparttable} 
}
\label{tab:results}
\end{table*}

\subsection{3.4~Implementation Details}
Table~\ref{tab:implementations} displays the main hyper-parameters of our experiments.
Unless otherwise specified, 
a) Different datasets often employ distinct pose data formats, such as COCO 18 for OU-MVLP, and BODY 25 for CCPG. To enhance flexibility, our implementation standardized these various formats to COCO 17 uniformly.
b) DeepGaitV2 denotes its pseudo-3D variant thanks to its computational efficiency. 
c) The double-side cutting strategy widely used for processing silhouettes is employed. The input size of skeleton maps is $2\times64\times44$. 
d) At the test phase, the entire sequence of skeleton maps will be directly fed into SkeletonGait and SkeletonGait++. As for the training stage, the data sampler collects a fixed-length segment of 30 frames as input. 
e) The spatial augmentation strategy suggested by~\cite{fan2022opengait} is adopted. 
f) The SGD optimizer with an initial learning rate of 0.1 and weight decay of 0.0005 is utilized.
g) The $\sigma$ controlling the variance in Eq.~\ref{equ:joints} and Eq.~\ref{equ:limbs} is set to 8.0 as default. 
h) Our code has been integrated into OpenGait~\cite{fan2022opengait}.

\section{4.~Experiments}
\label{sec:exp}

% \subsection{4.1~Datasets}
\noindent \textbf{Datasets}. 
Five popular gait datasets are employed for comprehensive comparisons, involving the OU-MVLP, SUSTech1K, CCPG, Gait3D, and GREW datasets.
Therefore, the comparison scope spans from fully constrained laboratories (the former three) to real-world scenarios (the latter two). 
Table~\ref{tab:datasets} shows the key statistical indicators.
Our experiments strictly follow the official evaluation protocols. 

\begin{table}[t]
\centering
\caption{Evaluation with different attributes on CCPG.}
    \scalebox{0.82}{%
    \setlength{\tabcolsep}{0.5em}

        \begin{tabular}{c|c|cc|cc|cc}
        
        \toprule
        \multirow{2}{*}{Input}   & \multirow{2}{*}{Method}     & \multicolumn{2}{c|}{CL-Full}      &  \multicolumn{2}{c|}{CL-UP} &  \multicolumn{2}{c}{CL-DN}  \\
          & & R-1 & mAP  & R-1 & mAP & R-1 & mAP                \\         \midrule[1pt]
        \multirow{2}{*}{\begin{tabular}[c]{@{}c@{}}Skeleton \\ Coordinates\end{tabular}} & GaitTR & 24.3 & 9.7 & 28.7 & 16.1 & 31.1 & 16.4 \\ 
        & GaitGraph2 & 5.0 & 2.4 & 5.7 & 4.0 & 7.3 & 4.2 \\ 
        \midrule
        Skeleton Maps & SkeletonGait & \textbf{52.4} & \textbf{20.8} & \textbf{65.4} & \textbf{35.8} & \textbf{72.8} & \textbf{40.3} \\ \midrule
        \multirow{5}{*}{Silhouette} & GaitSet & 77.7 & 46,4 & 83.5 & 59.6 & 83.2 & 61.4 \\ 
        & GaitPart & 77.8 & 45.5 & 84.5 & 63.1 & 83.3 & 60.1 \\ 
        & GaitGL & 69.1 & 27.0 & 75.0 & 37.1 & 77.6 & 37.6 \\ 
        & AUG-OGBase & 84.7 & 52.9 & 88.4 & 67.5 & 89.4 & 67.9 \\ 
        & DeepGaitV2     & \underline{\textbf{90.3}} & 62.0 & \underline{\textbf{96.3}} & \underline{\textbf{81.5}} & 91.5 & 78.1 \\ \midrule
        \multirow{2}{*}{\begin{tabular}[c]{@{}c@{}}Silhouette+\\ Skeleton\end{tabular}} & BiFusion & 77.5 & 46.7 & 84.8 & 64.1 & 84.8 & 61.9 \\ % \cmidrule{2-8}
        & SkeletonGait++ & 90.1 & \underline{\textbf{63.6}} & 95.4 & 81.1 & \underline{\textbf{92.5}} & \underline{\textbf{79.4}} \\ 
          \bottomrule
        \end{tabular}
        }
\label{tab:CCPG}
\end{table}

\begin{table*}[t]
\centering
\caption{Evaluation with different attributes on SUSTech1K.}
    \scalebox{0.83}{%
    \setlength{\tabcolsep}{0.5em}
        \begin{tabular}{c|cc|cccccccc|c|c}
        
        \toprule
        \multirow{2}{*}{Input} & \multirow{2}{*}{Method}   & \multirow{2}{*}{Publication}     & \multicolumn{8}{c}{Probe Sequence (R-1)}      &  \multicolumn{2}{|c}{Overall}                    \\
          & & & Normal    & Bag    & Clothing  & Carrying  & Umbrella  & Uniform  & Occlusion & Night   & R-1    & R-5  \\         \midrule
        \multirow{2}{*}{\begin{tabular}[c]{@{}c@{}}Skeleton \\ Coordinates\end{tabular}} & GaitTR      & Arxiv2022         & 33.3     & 31.5   & 21.0     & 30.4     & 22.7      & 34.6    & 44.9     & 23.5   & 30.8  & 56.0  \\
        & GaitGraph2    & CVPRW2022                                & 22.2     & 18.2   & 6.8     & 18.6     & 13.4      & 19.2    & 27.3     & 16.4   & 18.6  & 40.2  \\
        % & MSGG        & MTA2023                              & 67.11     & 66.16   & 35.92     & 63.31     & 61.58      & 58.07    & 66.59     & 17.88  &  33.8 & 82.82  \\ 
        \midrule
        Skeleton Maps & SkeletonGait    & Ours                             & \textbf{55.0}     & \textbf{51.0}   & \textbf{24.7}     & \textbf{49.9}     & \textbf{42.3}      & \textbf{52.0}    & \textbf{62.8}     & \textbf{43.9}   & \textbf{50.1}  & \textbf{72.6}   \\ \midrule
        \multirow{5}{*}{Silhouette} & GaitSet      & AAAI2019         & 69.1     & 68.2   & 37.4     & 65.0     & 63.1      & 61.0    & 67.2     & 23.0   & 65.0  & 84.8  \\
        & GaitPart    & CVPR2019                                & 62.2     & 62.8   & 33.1     & 59.5     & 57.2      & 54.8    & 57.2     & 21.7   & 59.2  & 80.8  \\
        & GaitGL        & ICCV2021                              & 67.1     & 66.2   & 35.9     & 63.3     & 61.6      & 58.1    & 66.6     & 17.9  &  63.1 & 82.8  \\ 
        & GaitBase    & CVPR2023                             & 81.5     & 77.5   & \underline{\textbf{49.6}}     & 75.8     & 75.5      & 76.7    & 81.4     & 25.9   & 76.1  & 89.4   \\ 
        & DeepGaitV2         & Arxiv2023                        & 83.5     & 79.5  & 46.3    & 76.8     & 79.1      & 78.5    & 81.1     & 27.3   & 77.4   & 90.2 \\ \midrule
        \multirow{2}{*}{\begin{tabular}[c]{@{}c@{}}Silhouette+\\ Skeleton\end{tabular}} & BiFusion      & MTA2023         & 69.8     & 62.3   & 45.4     & 60.9     & 54.3      & 63.5    & 77.8     & 33.7   & 62.1  & 83.4  \\
        & SkeletonGait++    & Ours                                & \underline{\textbf{85.1}}     & \underline{\textbf{82.9}}   & 46.6     & \underline{\textbf{81.9}}     & \underline{\textbf{80.8}}      & \underline{\textbf{82.5}}    & \underline{\textbf{86.2}}     & \underline{\textbf{47.5}}   & \underline{\textbf{81.3}}  & \underline{\textbf{95.5}}  \\
        \bottomrule
        \end{tabular}
        }
\label{tab:SUSTech1K}
\end{table*}

\begin{figure}[t]
\centering
\includegraphics[height=5.5cm]{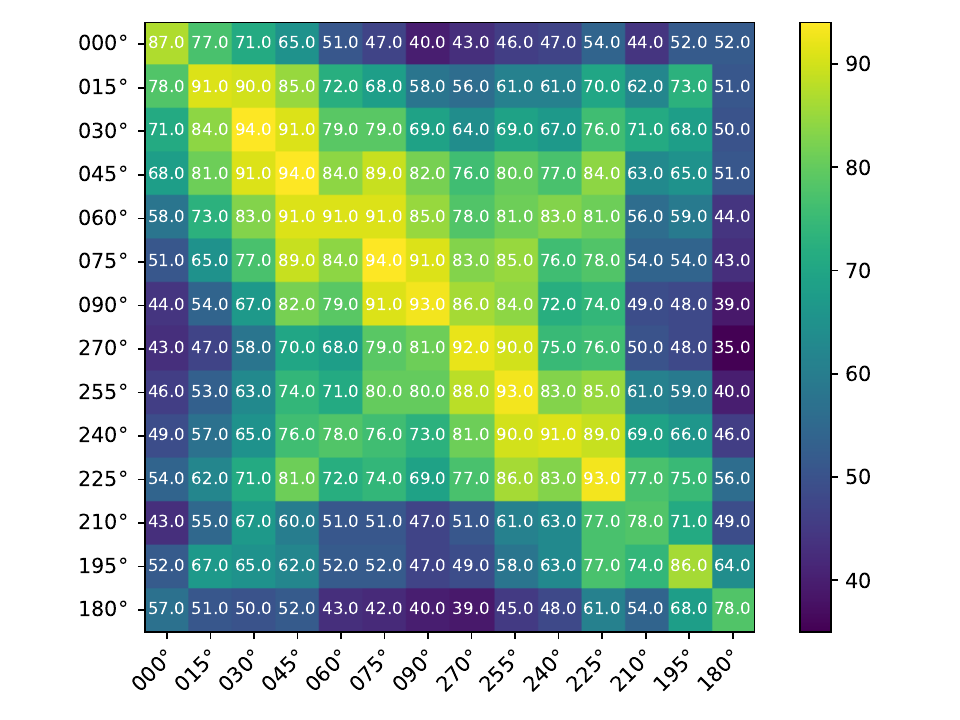}
\caption{SkeletonGait's performance on OU-MVLP over probe-gallery view pairs.}
\label{fig:viewpoint}
\end{figure}

\noindent \textbf{Compare SkeletonGait with Other Skeleton-based State-of-the-Arts}. 
As shown in Tab.~\ref{tab:results}, \ref{tab:CCPG}, and \ref{tab:SUSTech1K}, SkeletonGait outperforms the latest skeleton-based methods by breakthrough improvements in most cases. 
Specifically, it gains $+5.3\%$, $+22.9\%$, $+15.6\%$, $+36.5\%$, and $19.3\%$ (average/overall) rank-1 accuracy on the OU-MVLP, GREW, Gait3D, CCPG, and SUSTech1K datasets, respectively. 
To exclude the potential positive influence brought by the model size of SkeletonGait, we reduce its channels by half, thus making its model size nearly identical to that of GPGait, \textit{i.e.}, 2.85 v.s. 2.78M. 
After that, SkeletonGait reached the rank-1 accuracy of 33.2\% and 70.9\% on Gait3D and GREW, maintaining a higher performance than prior skeleton-based methods.

Since the skeleton map can be perceived as a silhouette that excludes body shape information, by comparing SkeletonGait with DeepGaitV2 in detail, we investigate that: 
\begin{itemize}
    \item \textbf{Importance}. 
    Structural features play a more important role than those shown by prior methods. 
    Or rather, it may contribute over 50\% according to the ratios between the performances of SkeletonGait and DeepGaitV2. 
    \item \textbf{Superiority}. 
    When silhouettes become relatively unreliable, \textit{e.g.}, the night case of SUSTech1K in Tab.~\ref{tab:SUSTech1K}, SkeletonGait surpasses DeepGaitV2 by a large margin, convincingly revealing the advantages of skeleton data. 
    \item \textbf{Challenge}. As shown in Fig.~\ref{fig:viewpoint}, the cross-view problem is still a major challenge for skeleton-based methods. 
    \item \textbf{Concerns about GREW}. The GREW dataset is widely acknowledged as the most challenging gait dataset due to its largest scale and real-world settings.
    However, SkeletonGait achieves a comparable performance compared to DeepGaitV2 on GREW, rather than on other relatively `easy' datasets. 
    In this paper, we observe that the gait pairs in GREW's test set seemly contain no many cross-view changes. 
    As mentioned, SkeletonGait works well on the cross-limited-view cases as shown in Fig.~\ref{fig:viewpoint}. 
    Therefore, we consider that the GREW dataset may lack viewpoint diversity, making its recognition task relatively easier compared with that of other datasets.
\end{itemize}

\begin{table}[t]
\centering
\caption{Ablation studies of SkeletonGait on Gait3D.}
\resizebox{0.80\columnwidth}{!}{
\begin{tabular}{c|cccc}
% \hline
\toprule
\multicolumn{1}{c|}{Control Variables} & \multicolumn{4}{c}{Recognition Performance} \\ \hline
$\sigma$                     & rank-1  & rank-5  & mAP  & mINP \\ \midrule % \hline
$\sigma=1.0$                      & 34.3    & 55.2    & 27   & 15.1 \\
% $\sigma=2.0$                      & 37.1    & \textbf{57.3}    & 28.6 & \textbf{16.0} \\
$\sigma=4.0$                      & 37.5    & \textbf{56.7}    & 28.5 & \textbf{16.1}  \\
$\sigma=8.0$                      & \textbf{38.1}    & \textbf{56.7}    & \textbf{28.9} & \textbf{16.1}  \\
$\sigma=16.0$                     & 36.0    & 55.2    & 26.9 & 15.0  \\ \bottomrule % \hline
\end{tabular}
}
\label{tab:ablation_skeletongait}
\end{table}

\begin{table}[t]
\centering
\caption{Ablation studies of SkeletonGait++ on Gait3D.}
\resizebox{1.0\columnwidth}{!}{
\begin{tabular}{c|ccc|ccc}
% \hline
\toprule
\multirow{2}{*}{Fusion Module} & \multicolumn{3}{c|}{Low-Level Fusion} & \multicolumn{3}{c}{High-Level Fusion} \\
                      & rank-1 & mAP  & mINP & rank-1  & mAP  & mINP \\ \midrule
Add                   & 76.5   & 69.6 & 41.9 & 76.2    & 69.5 & 41.7 \\
Concatenate           & 76.7   & 69.7 & 42.2 & 76.5    & 69.4 & 42.1 \\
Attention             & 77.6   & \textbf{70.3} & \textbf{42.6} & \textbf{78.2}    & 70.2 & 42.3 \\ \bottomrule
\end{tabular}
}
\label{tab:ablation_skeletongait++}
\end{table}

\begin{figure}[t]
\centering
\includegraphics[height=5.0cm]{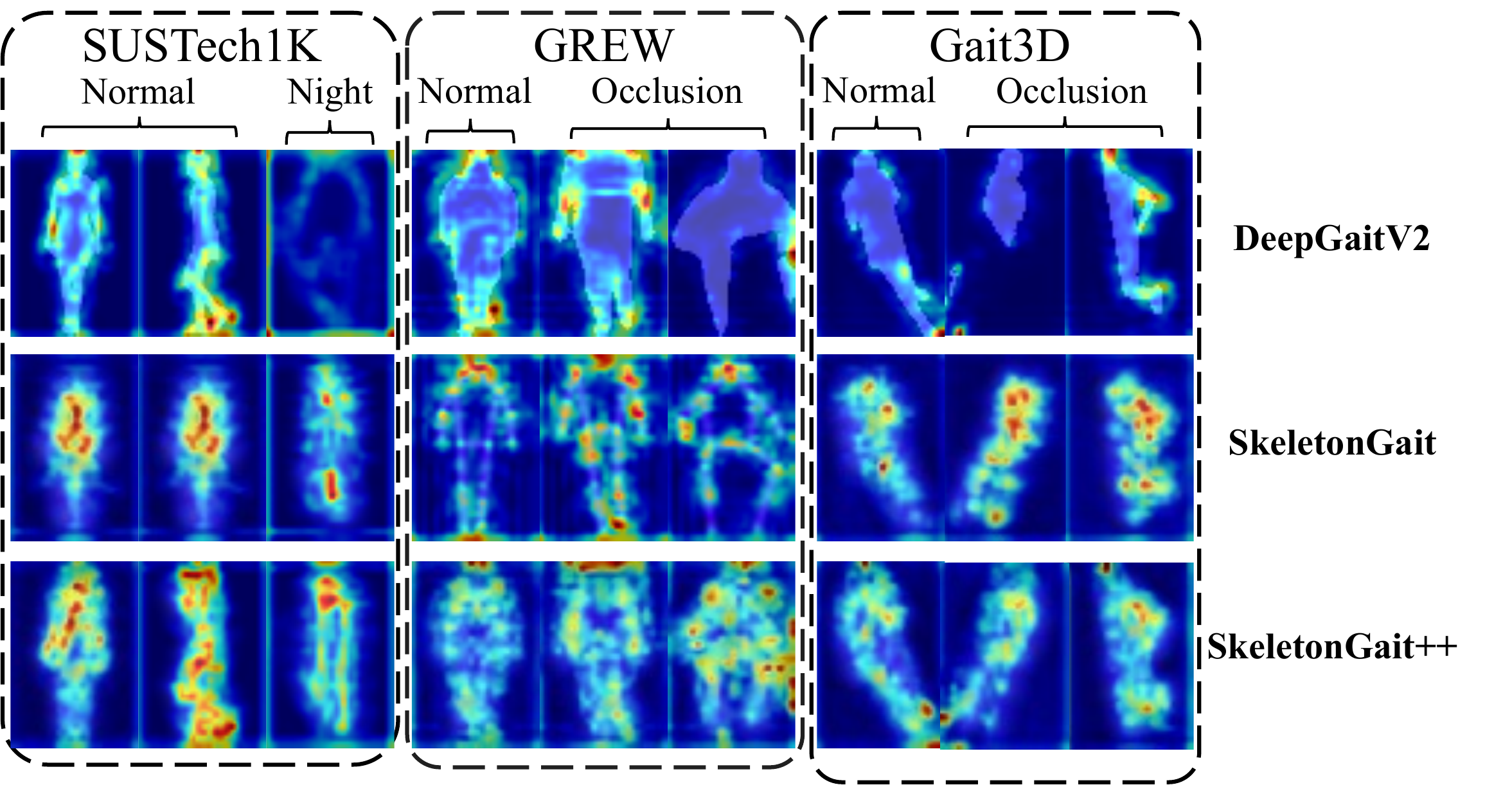}
\caption{The heatmaps~\cite{zhou2016learning} of DeepGaitV2 \textit{v.s.} SkeletonGait and SkeletonGait++. }
\label{fig:heatmap}
\end{figure}

\noindent \textbf{Compare SkeletonGait++ with Other State-of-the-Arts}. 
% Given that the silhouette and skeleton data of OU-MVLP do not align on a frame-by-frame basis, 
% we mainly evaluate SkeletonGait++ on GREW, Gait3D, CCPG, and SUSTech1K. 
According to Tab.~\ref{tab:results}, \ref{tab:CCPG}, and \ref{tab:SUSTech1K}, 
we find that: 
\begin{itemize}
    \item \textbf{Competitiveness}. 
    SkeletonGait++ reaches a new state-of-the-art with obvious gains, \textit{i.e.}, $+8.1\%$, $+3.2\%$, and $+5.2\%$ rank-1 accuracy on the GREW, Gait3D, and SUSTech1K, respectively. 
    As for the CCPG dataset, it also achieves overall superior performance. 
    \item \textbf{Benefits}. Compared to DeepGaitV2, the additional skeleton branch of SkeletonGait++ notably enhances the recognition accuracy, particularly when the body shape becomes less reliable. 
    This augmentation is particularly evident in challenging scenarios involving object carrying, occlusion, and poor illumination conditions, as observed on SUSTech1K dataset, \textit{i.e.}, Tab.~\ref{tab:SUSTech1K}.
    \item \textbf{Comprehensiveness}. 
    As shown in Fig.~\ref{fig:heatmap}, DeepGaitV2 directs its attention towards regions that exhibit distinct and discriminative body shapes. 
    On the other hand, SkeletonGait can only concentrate on `clean' structural features over the body joints and limbs.
    In comparison, SkeletonGait++ strikes a balance between these approaches, effectively capturing the `comprehensive' gait patterns that are rich in both body shape and structural characteristics. 
    Especially for night and occlusion cases, SkeletonGait++ adaptively leverages the still reliable skeleton branch to support the robust gait representation learning. This is an urgent need for practical applications, and we also think this is the main reason causing the performance gains on Gait3D and GREW datasets. 
\end{itemize}

Certainly, there are instances where skeleton data could also become unreliable, particularly in scenarios of extensive occlusion or other challenging conditions. 
However, experimental results reveal that skeleton data is more robust in such demanding situations than silhouette data on existing gait datasets.
This observation exhibits the significance of SkeletonGait++, as it effectively harnesses the strengths of both skeleton and silhouette data to tackle these challenges.

\noindent \textbf{Ablation Study}. 
% To determine the empirical optimal value of $\sigma$, we range it from 1 to 16 on the Gait3D dataset, and 
Table.~\ref{tab:ablation_skeletongait} shows that: 
a) SkeletonGait is robust to the value of $\sigma$.
b) $\sigma=8.0$ is an experimentally optimal choice.
% To determine the fusion location and mode, we try all the combinations on the Gait3D dataset, and 
Table.~\ref{tab:ablation_skeletongait++} reveals that: 
a) SkeletonGait++ is robust to both fusion location and mode. 
b) The low-level attention fusion is an experimentally optimal choice.

\section{5.~Discussions}
This paper introduces the skeleton map as a grid-based skeletal representation. 
The proposed SkeletonGait outperforms existing skeleton-based methods, emphasizing the importance of body structural features. 
SkeletonGait++ combines skeleton and silhouette features, achieving new state-of-the-art performance. 
The work demonstrates that model-based gait recognition has much to explore in the future.

\noindent \textbf{Acknowledgement}: 
This work was supported by the National Natural Science Foundation of China under Grant 61976144 and the Shenzhen International Research Cooperation Project under Grant GJHZ20220913142611021. 

\bibliography{aaai24}

\begin{thebibliography}{21}
\providecommand{\natexlab}[1]{#1}

\bibitem[{Chao et~al.(2019)Chao, He, Zhang, and Feng}]{chao2019gaitset}
Chao, H.; He, Y.; Zhang, J.; and Feng, J. 2019.
\newblock Gaitset: Regarding gait as a set for cross-view gait recognition.
\newblock In \emph{Proceedings of the AAAI conference on artificial
  intelligence}, volume~33, 8126--8133.

\bibitem[{Duan et~al.(2022)Duan, Zhao, Chen, Lin, and Dai}]{duan2022revisiting}
Duan, H.; Zhao, Y.; Chen, K.; Lin, D.; and Dai, B. 2022.
\newblock Revisiting skeleton-based action recognition.
\newblock In \emph{Proceedings of the IEEE/CVF Conference on Computer Vision
  and Pattern Recognition}, 2969--2978.

\bibitem[{Fan et~al.(2023)Fan, Hou, Huang, and Yu}]{fan2023exploring}
Fan, C.; Hou, S.; Huang, Y.; and Yu, S. 2023.
\newblock Exploring Deep Models for Practical Gait Recognition.
\newblock \emph{arXiv preprint arXiv:2303.03301}.

\bibitem[{Fan et~al.(2022)Fan, Liang, Shen, Hou, Huang, and
  Yu}]{fan2022opengait}
Fan, C.; Liang, J.; Shen, C.; Hou, S.; Huang, Y.; and Yu, S. 2022.
\newblock {OpenGait}: Revisiting Gait Recognition Toward Better Practicality.
\newblock \emph{arXiv preprint arXiv:2211.06597}.

\bibitem[{Fan et~al.(2020)Fan, Peng, Cao, Liu, Hou, Chi, Huang, Li, and
  He}]{fan2020gaitpart}
Fan, C.; Peng, Y.; Cao, C.; Liu, X.; Hou, S.; Chi, J.; Huang, Y.; Li, Q.; and
  He, Z. 2020.
\newblock Gaitpart: Temporal part-based model for gait recognition.
\newblock In \emph{Proceedings of the IEEE/CVF conference on computer vision
  and pattern recognition}, 14225--14233.

\bibitem[{Li et~al.(2023)Li, Hou, Zhang, Cao, Liu, Huang, and
  Zhao}]{Li_2023_CVPR}
Li, W.; Hou, S.; Zhang, C.; Cao, C.; Liu, X.; Huang, Y.; and Zhao, Y. 2023.
\newblock An In-Depth Exploration of Person Re-Identification and Gait
  Recognition in Cloth-Changing Conditions.
\newblock In \emph{Proceedings of the IEEE/CVF Conference on Computer Vision
  and Pattern Recognition (CVPR)}, 13824--13833.

\bibitem[{Li et~al.(2020)Li, Makihara, Xu, Yagi, Yu, and Ren}]{li2020end}
Li, X.; Makihara, Y.; Xu, C.; Yagi, Y.; Yu, S.; and Ren, M. 2020.
\newblock End-to-end model-based gait recognition.
\newblock In \emph{Proceedings of the Asian conference on computer vision}.

\bibitem[{Liao et~al.(2022)Liao, Li, Bhattacharyya, and
  York}]{liao2022posemapgait}
Liao, R.; Li, Z.; Bhattacharyya, S.~S.; and York, G. 2022.
\newblock PoseMapGait: A model-based gait recognition method with pose
  estimation maps and graph convolutional networks.
\newblock \emph{Neurocomputing}, 501: 514--528.

\bibitem[{Liao et~al.(2020)Liao, Yu, An, and Huang}]{liao2020model}
Liao, R.; Yu, S.; An, W.; and Huang, Y. 2020.
\newblock A model-based gait recognition method with body pose and human prior
  knowledge.
\newblock \emph{Pattern Recognition}, 98: 107069.

\bibitem[{Lin, Zhang, and Yu(2021)}]{lin2021gait}
Lin, B.; Zhang, S.; and Yu, X. 2021.
\newblock Gait recognition via effective global-local feature representation
  and local temporal aggregation.
\newblock In \emph{Proceedings of the IEEE/CVF International Conference on
  Computer Vision}, 14648--14656.

\bibitem[{Liu and Yuan(2018)}]{liu2018recognizing}
Liu, M.; and Yuan, J. 2018.
\newblock Recognizing human actions as the evolution of pose estimation maps.
\newblock In \emph{Proceedings of the IEEE conference on computer vision and
  pattern recognition}, 1159--1168.

\bibitem[{Nixon and Carter(2006)}]{nixon2006automatic}
Nixon, M.~S.; and Carter, J.~N. 2006.
\newblock Automatic recognition by gait.
\newblock \emph{Proceedings of the IEEE}, 94(11): 2013--2024.

\bibitem[{Peng et~al.(2023)Peng, Ma, Zhang, and He}]{peng2023learning}
Peng, Y.; Ma, K.; Zhang, Y.; and He, Z. 2023.
\newblock Learning rich features for gait recognition by integrating skeletons
  and silhouettes.
\newblock \emph{Multimedia Tools and Applications}, 1--22.

\bibitem[{Shen et~al.(2023)Shen, Fan, Wu, Wang, Huang, and Yu}]{Shen_2023_CVPR}
Shen, C.; Fan, C.; Wu, W.; Wang, R.; Huang, G.~Q.; and Yu, S. 2023.
\newblock LidarGait: Benchmarking 3D Gait Recognition With Point Clouds.
\newblock In \emph{Proceedings of the IEEE/CVF Conference on Computer Vision
  and Pattern Recognition (CVPR)}, 1054--1063.

\bibitem[{Shen et~al.(2022)Shen, Yu, Wang, Huang, and
  Wang}]{shen2022comprehensive}
Shen, C.; Yu, S.; Wang, J.; Huang, G.~Q.; and Wang, L. 2022.
\newblock A comprehensive survey on deep gait recognition: algorithms, datasets
  and challenges.
\newblock \emph{arXiv preprint arXiv:2206.13732}.

\bibitem[{Takemura et~al.(2018)Takemura, Makihara, Muramatsu, Echigo, and
  Yagi}]{takemura2018multi}
Takemura, N.; Makihara, Y.; Muramatsu, D.; Echigo, T.; and Yagi, Y. 2018.
\newblock Multi-view large population gait dataset and its performance
  evaluation for cross-view gait recognition.
\newblock \emph{IPSJ Transactions on Computer Vision and Applications}, 10(1):
  1--14.

\bibitem[{Teepe et~al.(2021)Teepe, Khan, Gilg, Herzog, H{\"o}rmann, and
  Rigoll}]{teepe2021gaitgraph}
Teepe, T.; Khan, A.; Gilg, J.; Herzog, F.; H{\"o}rmann, S.; and Rigoll, G.
  2021.
\newblock Gaitgraph: Graph convolutional network for skeleton-based gait
  recognition.
\newblock In \emph{2021 IEEE International Conference on Image Processing
  (ICIP)}, 2314--2318. IEEE.

\bibitem[{Wang et~al.(2022)Wang, Zhang, Shen, Du, Zhao, Cui, and Wen}]{9337225}
Wang, Y.; Zhang, X.; Shen, Y.; Du, B.; Zhao, G.; Cui, L.; and Wen, H. 2022.
\newblock Event-Stream Representation for Human Gaits Identification Using Deep
  Neural Networks.
\newblock \emph{IEEE Transactions on Pattern Analysis and Machine
  Intelligence}, 44(7): 3436--3449.

\bibitem[{Zheng et~al.(2022)Zheng, Liu, Liu, He, Yan, and Mei}]{zheng2022gait}
Zheng, J.; Liu, X.; Liu, W.; He, L.; Yan, C.; and Mei, T. 2022.
\newblock Gait Recognition in the Wild with Dense 3D Representations and A
  Benchmark.
\newblock In \emph{Proceedings of the IEEE/CVF Conference on Computer Vision
  and Pattern Recognition}, 20228--20237.

\bibitem[{Zhou et~al.(2016)Zhou, Khosla, Lapedriza, Oliva, and
  Torralba}]{zhou2016learning}
Zhou, B.; Khosla, A.; Lapedriza, A.; Oliva, A.; and Torralba, A. 2016.
\newblock Learning deep features for discriminative localization.
\newblock In \emph{Proceedings of the IEEE conference on computer vision and
  pattern recognition}, 2921--2929.

\bibitem[{Zhu et~al.(2021)Zhu, Guo, Yang, Huang, Deng, Huang, Du, Lu, and
  Zhou}]{zhu2021gait}
Zhu, Z.; Guo, X.; Yang, T.; Huang, J.; Deng, J.; Huang, G.; Du, D.; Lu, J.; and
  Zhou, J. 2021.
\newblock Gait recognition in the wild: A benchmark.
\newblock In \emph{Proceedings of the IEEE/CVF international conference on
  computer vision}, 14789--14799.

\end{thebibliography}

\end{document}